%% file: main.tex
\documentclass[journal,twoside,web]{ieeecolor}
\usepackage{jsen}
\usepackage{cite}
\usepackage{amsmath,amssymb,amsfonts}
\usepackage{algorithmic}
\usepackage{graphicx}
\usepackage{textcomp}
\usepackage{xcolor}
\usepackage{wrapfig}
\usepackage{multirow}
\usepackage[acronym]{glossaries}
\usepackage{booktabs}
\let\labelindent\relax
\usepackage[inline]{enumitem}
\usepackage{siunitx}
\usepackage{overpic}
\usepackage{hyperref}
\usepackage{xcolor, tikz}
\usepackage[normalem]{ulem}
\newcommand{\revdel}[1]{}

\hypersetup{
    colorlinks=false, % Colored links instead of boxed
    urlcolor=black,  % Color of external links
    pdfborder={0 0 0} % Remove border around links
}
\usepackage[capitalise]{cleveref}

\def\revcolor{black}

\newcommand{\revone}[1]{\textcolor{\revcolor}{#1}}
\usepackage{eso-pic}
\usepackage{url}
\AddToShipoutPictureBG*{
  \AtPageUpperLeft{%
    \put(0,-40){\raisebox{15pt}{\makebox[\paperwidth]{\begin{minipage}{21cm}\centering
      \textcolor{gray}{This paper has been accepted for publication in the IEEE Sensors Journal.}
    \end{minipage}}}}%
  }
  \AtPageLowerLeft{%
    \raisebox{20pt}{\makebox[\paperwidth]{\begin{minipage}{21cm}\centering
      \textcolor{gray}{ \copyright 2024 IEEE.  Personal use of this material is permitted.  Permission from IEEE must be obtained for all other uses, in any current or future media, including reprinting/republishing this material for advertising or promotional purposes, creating new collective works, for resale or redistribution to servers or lists, or reuse of any copyrighted component of this work in other works.
      }
    \end{minipage}}}%
  }
}
\def\BibTeX{{\rm B\kern-.05em{\sc i\kern-.025em b}\kern-.08em
    T\kern-.1667em\lower.7ex\hbox{E}\kern-.125emX}}
\definecolor{abstractbg}{rgb}{0.89804,0.94510,0.83137}
\begin{document}
\bstctlcite{IEEEexample:BSTcontrol} %used tp shortn author list (needs some text in the bib file too)
\include{abr}
\title{DSORT-MCU: Detecting Small Objects in Real-Time on Microcontroller Units}
\author{Liam Boyle\textsuperscript{1}, Julian Moosmann\textsuperscript{1}, Nicolas Baumann\textsuperscript{1}, Dr. Seonyeong Heo\textsuperscript{2}, Dr. Michele Magno\textsuperscript{1} 
\thanks{Submitted for review on the 29\textsuperscript{th} of February 2024 }
\thanks{\textsuperscript{1} Dept. of Information Technology and Electrical Engineering, ETH Zurich, 8092 Zurich, Switzerland}
\thanks{\textsuperscript{2} Dept. of Computer Science and Engineering, Kyung Hee University, 130-701 Yongin, South Korea}
}
% \thanks
% {This paragraph of the first footnote will contain the date on 
% which you submitted your paper for review. It will also contain support 
% information, including sponsor and financial support acknowledgment. For 
% example, ``This work was supported in part by the U.S. Department of 
% Commerce under Grant BS123456.'' }
% \thanks{The next few paragraphs should contain 
% the authors' current affiliations, including current address and e-mail. For 
% example, F. A. Author is with the National Institute of Standards and 
% Technology, Boulder, CO 80305 USA (e-mail: author@boulder.nist.gov). }
% \thanks{S. B. Author, Jr., was with Rice University, Houston, TX 77005 USA. He is 
% now with the Department of Physics, Colorado State University, Fort Collins, 
% CO 80523 USA (e-mail: author@lamar.colostate.edu).}
% \thanks{T. C. Author is with 
% the Electrical Engineering Department, University of Colorado, Boulder, CO 
% 80309 USA, on leave from the National Research Institute for Metals, 
% Tsukuba, Japan (e-mail: author@nrim.go.jp).
% }
% }

\hyphenation{Tiny-issimo-YOLO}

\IEEEtitleabstractindextext{%
\fcolorbox{abstractbg}{abstractbg}{%
\begin{minipage}{\textwidth}%
\begin{wrapfigure}[19]{r}{3.2in}%
\includegraphics[width=3in]{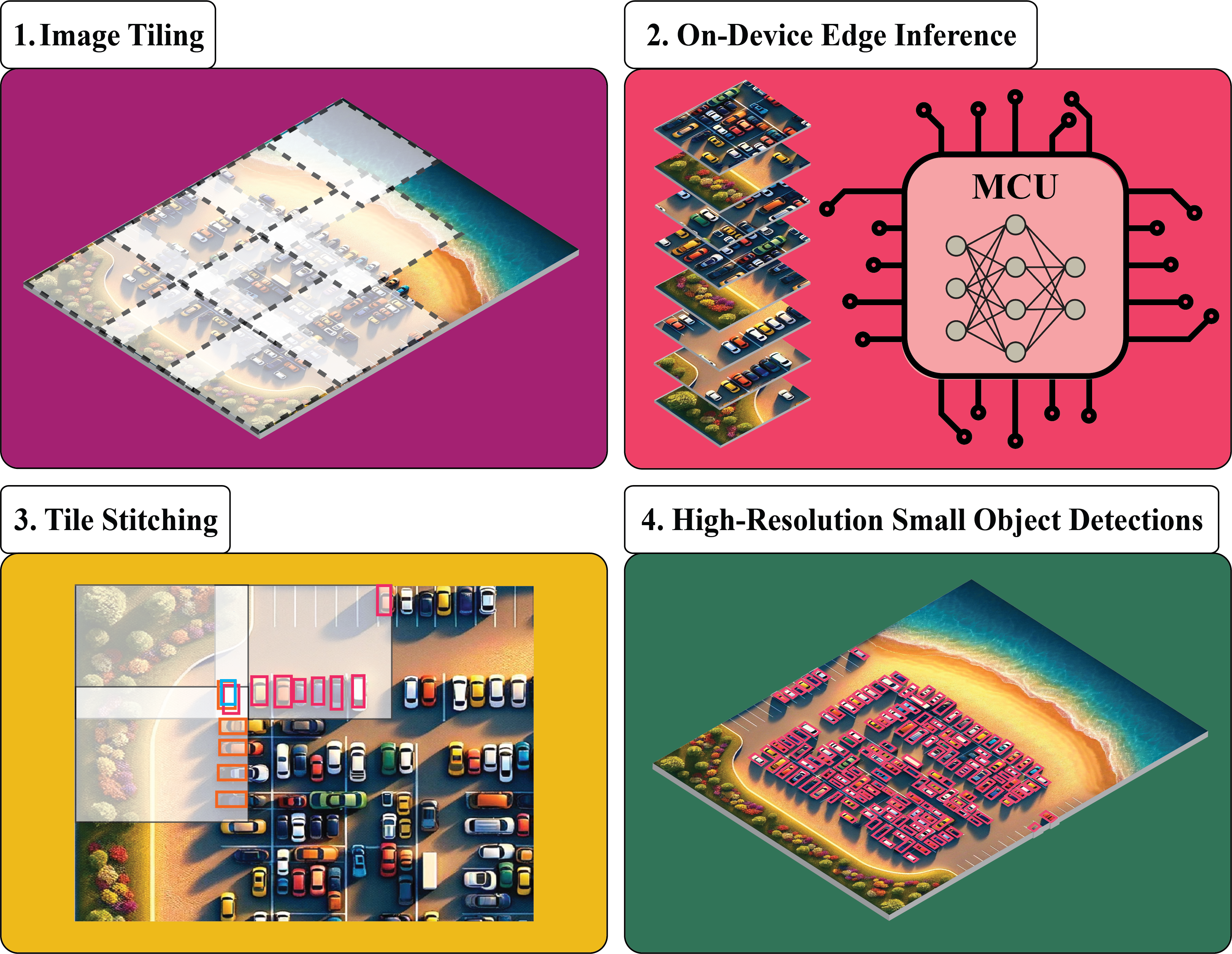}%
% \begin{overpic}[width=3.2in]{figures/fomo_arch_cars_screenshot_bold_transparent.png}
%         \put(90,-5){\textcolor{black}{{\tiny \cite{liam_tiling}}}}
% \end{overpic}
\end{wrapfigure}%
\begin{abstract}
Advances in lightweight neural networks have revolutionized computer vision in a broad range of \gls{iot} applications, encompassing remote monitoring and process automation. However, the detection of small objects, which is crucial for many of these applications, remains an underexplored area in current computer vision research, particularly for low-power embedded devices that host resource-constrained processors. To address said gap, this paper proposes an adaptive tiling method for lightweight and energy-efficient object detection networks, including YOLO-based models and the popular \gls{fomo} network. The proposed tiling enables object detection on low-power \glspl{mcu} with no compromise on accuracy compared to large-scale detection models. The benefit of the proposed method is demonstrated by applying it to \gls{fomo} and TinyissimoYOLO networks on a novel\revdel{RISC-V} \revone{\emph{RISC-V}}-based \gls{mcu} with built-in \gls{ml} accelerators. Extensive experimental results show that the proposed tiling method boosts the F1-score by up to 225\% for both \gls{fomo} and TinyissimoYOLO networks while reducing the average object count error by up to 76\% with \gls{fomo} and up to 89\% for TinyissimoYOLO. Furthermore, the findings of this work indicate that using a soft F1 loss over the popular binary cross-entropy loss can serve as an implicit non-maximum suppression for the \gls{fomo} network.\revdel{To evaluate the real-world performance the networks are}\revdel{deployed on the RISC-V based GAP9 microcontroller from Greenwaves' Technology} \revone{To evaluate the real-world performance, the networks are deployed on the \emph{RISC-V} based GAP9 microcontroller from \emph{GreenWaves Technologies}}, showcasing the proposed method's ability to strike a balance between detection performance ($58\% - 95\%$ F1 score), low latency (\SI{0.6}{\milli\second/Inference} - \SI{16.2}{\milli\second/Inference}), and energy efficiency (\SI{31}{\micro\joule/Inference} - \SI{1.27}{\milli\joule/Inference}) while performing multiple predictions using high-resolution images on a \gls{mcu}.
% \newline
\end{abstract}
\glsreset{iot}
\glsreset{ml}
\glsreset{fomo}
\glsreset{mcu}
% \todo[inline]{Update abstract. How is this typically done for paper extensions?}
\end{minipage}}}

\maketitle

\section{Introduction}
\label{sec:introduction}
%\IEEEPARstart{T}{he} introduction of low-power \gls{iot} devices equipped with sensors and novel \gls{ml} algorithms is revolutionizing various domains in the realm of health monitoring \cite{wang2022mi, moin2021wearable}, home \cite{popa2019deep} and process automation \cite{kamal2022architectural}. These edge devices enhance privacy, optimize bandwidth utilization, and reduce costs by running on-device \gls{ml} algorithms to extract non-critical semantic information that can be sent to the cloud. % Many of these tasks rely on accurate object detection, often of very small objects size, which remains as a challenge on memory-bound devices such as \glspl{mcu}. While high-resolution images are preferred for detecting small-sized objects\cite{}, RGB images with 8-bit color depth and a resolution of 577 by 577 already exceeds \SI{1}{\mega\byte} of flash memory required for storage. Utilizing this paper's previously proposed image tiling method to decrease on-device image memory requirements while further increasing detection accuracy with TinyissimoYOLO network. This paper extends the results previously obtained by applying the tiling method to \gls{fomo} and TinyissimoYOLO networks and deploying them on GAP9.

%Many of these use-cases(?) rely on accurate object detection. Today, state-of-the-art object \cite{yolo, rcnn, ssd} detectors typically use \gls{cnn} architectures to predict object locations and classes. More recently, transformer-based network architectures \cite{vit} have gained considerable popularity.

\IEEEPARstart{T}{he} integration of low-power \gls{iot} devices, equipped with advanced sensors and cutting-edge \gls{ml} algorithms, is driving changes across various sectors, including health monitoring \cite{moin2021wearable}, home automation \cite{popa2019deep}, and industrial process optimization \cite{kamal2022architectural}. These edge computing devices increase data privacy, enhance bandwidth efficiency, and contribute to cost reductions by leveraging on-device \gls{ml} algorithms. These algorithms process privacy-critical sensory information locally and extract metadata, which is subsequently transmitted to the cloud for further analysis or action. A critical application for many of these scenarios is precise object detection \cite{sonkar2022real,yang2022vamyolox}. Presently, state-of-the-art object detectors \cite{song2022ms, vit, ssd, fpn} typically use \gls{cnn} architectures to predict object locations and classes within an image. In a noteworthy shift, transformer-based network architectures \cite{detr, vit} have emerged as a powerful alternative, demonstrating significant promise in advancing the capabilities of object detection systems. While these networks have deep contextual understanding, they are extremely computationally demanding, and thus require powerful and expensive hardware consuming several hundred watts, making them unsuitable for the majority of \gls{iot} processors \cite{saha2022machine}. %eggimann2019risc
This has prompted scholars to propose new lightweight network architectures utilizing model quantization methods to enable \gls{tinyml} on mobile and low-power embedded devices \cite{moosmann2023tinyissimoyolo, ancilotto2023xinet, mcunet, lin2021mcunetv2}. %\cite{acurate_detection_edge, info_flow}
While advances in lightweight neural networks have allowed great progress to be made on tasks such as image classification \cite{saha2022machine}, and recently started to yield good detection accuracy on simplified object detection tasks \cite{fomo, mcunet, moosmann2023tinyissimoyolo, ancilotto2023xinet, paissan2022phinets}, detecting small objects remains a challenge \cite{thinNets, smallod0}, especially on memory-bound devices such as \glspl{mcu}. The limited computational budget on these embedded processors \cite{lin2021mcunetv2, moosmann2023flexible} %\cite{liu2024joint}
results in high execution latency of multiple seconds for one image, as was shown in the prior work of this paper\revdel{$^1$} presented at IEEE Sensors \cite{liam_tiling}.

%The challenge of detecting small objects in images is primarily attributed to the limited pixel coverage these objects have, which significantly inhibits the ability of object detectors to generate distinctive features, a crucial step for accurate identification  \cite{liam_tiling}. This limitation is fundamentally due to the small objects occupying only a tiny fraction of the image's overall pixel array, resulting in insufficient detail for traditional detection algorithms to effectively differentiate these objects from the surrounding environment. High-resolution images increase the signal-to-noise ratio, however, RGB images with 8-bit color depth and a resolution of $577\times 577$ already exceed the \SI{1}{\mega\byte} of flash memory typically available on low power \glspl{mcu}, further increasing the challenge for small object detectors on memory constrained devices typically used for \gls{tinyml} algorithms \cite{saha2022machine}. Another factor is the striding operations in \gls{cnn}-based algorithms resulting in small objects disappearing completely in the feature maps. Furthermore, it is worth noting that most of the publicly available datasets, such as ImageNet \cite{imagenet}, display a bias towards large objects (i.e. prominently featured objects). This adversely affects the usefulness of transfer learning which is an important tool for developing models for specific use cases with limited datasets. 

The challenge of detecting small objects in images is primarily attributed to the limited pixel coverage these objects have, which significantly inhibits the ability of object detectors to generate distinctive features, a crucial step for accurate identification  \cite{liam_tiling}. High-resolution images increase the signal-to-noise ratio, however, RGB images with 8-bit color depth and a resolution of $577\times 577$ \revone{pixels} already exceed the \SI{1}{\mega\byte} of flash memory typically available on low power \glspl{mcu}. The limited remaining memory restricts the size of models that can be deployed \cite{saha2022machine}. Another factor is the striding operations in \gls{cnn}-based algorithms resulting in small objects disappearing completely in the feature maps.

% \begin{figure}
% \centering
% \includegraphics[width=0.45\textwidth]{figures/fomo_arch_cars_screenshot_bold.png}
% \caption{Overview of the \gls{fomo} architecture. We show filter sizes for an example input image size of $224 \times 224 \times 3$. In this example, the feature extractor produces a $28 \times 28$ feature grid for which the detection head classifies each grid cell as containing an object or not. The final output after clustering is shown on the right.}
% \label{fig:fomo-arch}
% \vspace{-10pt}
% \end{figure}
\begin{figure}
\centering
\includegraphics[width=0.5\textwidth]{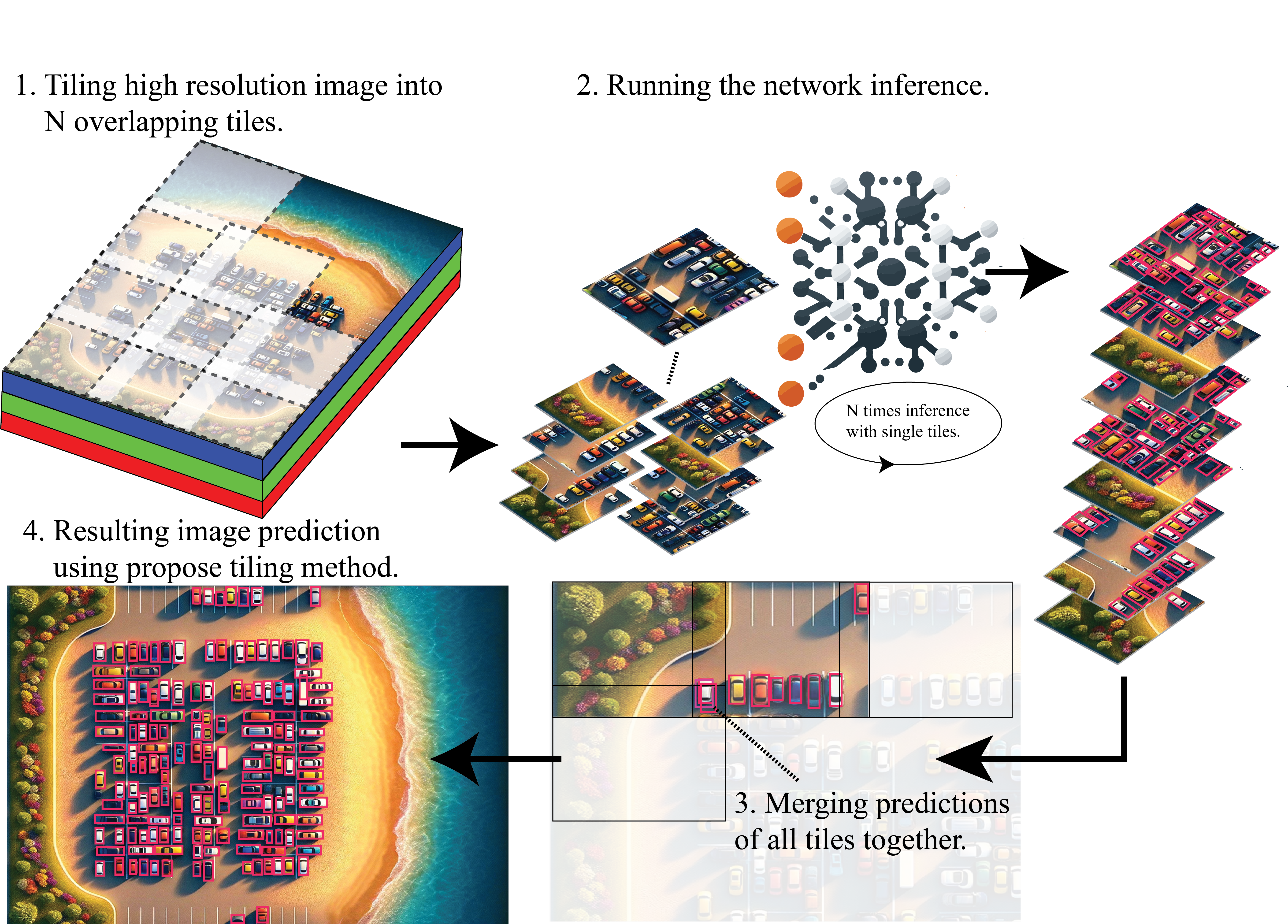}
\caption{Overview of the proposed adaptive tiling method for small object detection.}
\label{fig:tiling_overview}
\vspace{-10pt}
\end{figure}

%This paper is an extension of the work presented in \textit{"Enhancing Lightweight Neural Networks for Small Object Detection in IoT Applications"}\footnote{https://doi.org/10.1109/SENSORS56945.2023.10325126}, where an adaptive tiling method was proposed to enable the detection of small objects with lightweight networks like \gls{fomo}. By tiling images and deploying the network to the Sony \textit{Spresense} \cite{spresense}, it was shown that good detection results are possible on a low-power \gls{mcu}, though a high latency penalty is incurred. Furthermore, it was seen in the previous work \cite{liam_tiling} that \gls{fomo} reaches a limit in its capacity to predict objects, where further increasing the tiling and input resolution no longer improves the detection performance.\newline

This paper serves as a comprehensive elaboration and extension of the methodologies and findings introduced in our prior work \textit{"Enhancing Lightweight Neural Networks for Small Object Detection in IoT Applications\footnotemark[1]"}. The cornerstone of our previous study was the development of an adaptive tiling technique---see \cref{fig:tiling_overview} for an overview of the proposed approach---designed specifically to augment the capabilities of lightweight neural networks, such as \gls{fomo} \cite{fomo}, in accurately identifying small objects. This technique's efficacy is empirically demonstrated through its implementation on the\revdel{Sony Spresense} \revone{\emph{Sony Spresense}} \gls{mcu}, a platform characterized by its low power consumption. Despite achieving commendable detection accuracy, the approach was constrained by a notable latency penalty. Additionally, our previous investigations revealed an inherent limitation in \gls{fomo}'s object prediction capacity, where enhancements in tiling granularity and input resolution ceased to yield proportional improvements in detection performance.

Building upon these foundational insights, this extension of our previous work introduces significant advancements and novel contributions, encapsulated in the following aspects:
\begin{itemize}
\item \textbf{Enhanced Methodological Depth}: This paper extension provides an in-depth explanation of the adaptive tiling methodology introduced in \cite{liam_tiling}, showcasing its ability to substantially increase the signal-to-noise ratio without necessitating a corresponding increase in network input resolution. %This detailed exploration elucidates the method's underpinning principles and its strategic advantage in small object detection tasks.

\item \textbf{Network Modifications and Integrations}: This work describes in more detail the modifications applied to the original \gls{fomo} architecture and their synergistic effects when combined with our adaptive tiling approach. An ablation study is performed to demonstrate their role in enhancing detection capabilities beyond the baseline established in our previous research.

\end{itemize}

Expanding beyond the scope of our initial publication, this extension broadens the empirical validation of our approach through the following experimental advancements:
\begin{itemize}
\item  \textbf{Application to TinyissimoYOLO}: By applying the adaptive tiling technique to the TinyissimoYOLO network, state-of-the-art detection accuracies are achieved compared to the latest benchmarks set on the CARPK \cite{carpk} dataset. This accomplishment underscores the versatility and effectiveness of our method across different network architectures.

\item  \textbf{Deployment on GAP9}: Both the \gls{fomo} and TinyissimoYOLO networks are deployed on the GAP9 \gls{mcu}, a novel\revdel{RISC-V} \revone{\emph{RISC-V}}-based parallel processor with a hardware accelerator. This platform enables state-of-the-art inference latency, marking a significant milestone in the pursuit of efficient, real-time object detection in IoT applications.

\item \textbf{Open Source Code\footnote{\url{https://github.com/ETH-PBL/TinyissimoYOLO}}}: An open source version of the proposed tiling method is released together with TinyissimoYOLO.
\end{itemize}
Through these contributions, this paper extension not only extends the theoretical and practical understanding of small object detection within the IoT paradigm but also sets a new benchmark in the deployment of neural networks on low-power \glspl{mcu}, thereby paving the way for future innovations in the field.

\section{Related Work}
The detection of small objects remains a challenging area of research owing to the inherent difficulty in generating features from the relatively limited quantity of pixels that cover small objects. The majority of widely-used object detection models \cite{yolov4, mobilenetsv2, fasterrcnn, detr, efficientdet} are not optimized to deal with this challenge. \Cref{tab:small_object_det} highlights popular object detectors like YOLOv4 \cite{yolov4} and Faster R-CNN \cite{fasterrcnn} performing significantly worse on small objects. This limitation can be attributed, in part, to the network architecture. For example, YOLO \cite{yolo} utilizes a grid-based approach to make predictions, whereby small objects that fall within the same grid cell may not be detected correctly. An additional issue arises in relation to the training data used to develop these models. The majority of object detection models require a substantial volume of training data, posing a challenge for niche use cases where only limited data is available. As such, practitioners may rely on pre-trained models that can be fine-tuned with data specific to their use case. However, the general-purpose training datasets utilized to pre-train models are often biased towards objects that occupy significant portions of the image \cite{small_object_dataset}. Thus, limiting their usefulness as datasets for pre-training when the target use case contains mainly small objects.

\begin{table}
\centering
\caption[Common Object Detectors Performance on Small Objects]{Accuracy comparison of several object detectors on the MS COCO dataset \cite{coco}. All shown methods achieve significantly lower average precision on small objects, indicated by $AP_S$.}
\begin{tabular}{llll}
\toprule
\toprule
Method        & $AP_S$ [\%]    & $AP_M$ [\%]    & $AP_L$ [\%]     \\
\midrule
\midrule
Faster R-CNN \cite{fasterrcnn}  & 21,8          & 42,6          & 50,7           \\ 
% \hline
YOLOv4 \cite{yolov4}       & \textbf{24,3} & \textbf{46,1} & \textbf{55,2}  \\ 
% \hline
EfficientDet \cite{efficientdet} & 12,0          & 38,3          & 51,2         \\
\bottomrule
\bottomrule
\end{tabular}

\label{tab:small_object_det}
\end{table}

% general object detection related work
Efforts made to tackle these issues can broadly be categorized as either trying to increase the amount of information available for a given object or as trying to maximize the expressiveness of the features generated on small objects. The work by Liu et al. \cite{ssd} was one of the first to take advantage of \gls{cnn}s pyramidal structure to make predictions at multiple scales. They add multiple feature layers, with progressively smaller resolutions, at the end of their network. Each of these feature layers makes its own predictions, and finally, a non-maximum suppression step chooses the most likely predictions. However, the downsampling that occurs in the initial feature extractor limits the capacity of this network to detect small objects. This limitation is addressed in the work of Lin et al. \cite{fpn}, who propose to combine the high-resolution features from early layers with the more mature features from later layers. To achieve this, they concatenate feature maps from multiple resolutions and employ a single predictor that accepts this multi-scale feature vector as input. These findings inform the decision of this work to employ TinyissimoYOLOv1.3 \cite{moosmann2023ultra}, a lightweight network that uses the detection head introduced by YOLOv3 \cite{yolov3}, which combines features from different scales.

% small object detection related work
While these approaches try to maximize the expressiveness of features for small objects through changes in the network architecture, other researchers have tried to achieve this goal by increasing the amount of information available to create features. An obvious approach is to increase the input resolution of the network, which is, suggested in the work of Liu et al. \cite{ssd}. Unfortunately, increasing the image resolution is not always possible. Many networks have fixed input sizes, and increasing the image size greatly increases the memory and compute load. Cagatay et al. \cite{sahi} propose a framework called Slicing Aided Hyper Inference for tiling images into sub-images. They show how this framework can be used to improve the inference performance for off-the-shelf object detectors while achieving better results when the framework is used to fine-tune pre-trained models. Their method divides an input image into a fixed number of tiles and combines predictions from each tile with predictions made on the full image in a post-processing step. 
%The drawback of dividing images into tiles for individual processing is an increase in total inference time, which can be problematic for applications that depend on real-time inference.
To offset the increased processing time introduced by the image tiling Plastiras et al. \cite{selective_tiling} propose a selective tiling approach. Their approach assigns each tile an importance score, and for a given iteration it only processes the tiles with the highest scores. Inspired by the selective tiling approach, the previous work introduces an \textit{adaptive} tiling approach \cite{liam_tiling} to minimize the number of inference passes while optimizing object size to ensure maximal detection accuracy.

%this work introduces an adaptive tiling approach to minimize the number of inference passes while optimizing object size to ensure best detection accuracy.

% object detection on \gls{mcu}
Deployment of \glspl{cnn} on \gls{mcu} class devices are mostly memory-bound problems requiring the approaches to deal with \begin{enumerate*}[label=(\roman*),,font=\itshape]
\item quantization techniques \cite{han-pruning}, to decrease \gls{ram} and FLASH utilization while trying to keep accuracy constant \cite{liang2021pruning}, 
\item reducing the network size, such as the approach of \gls{fomo} \cite{fomo} or TinyissimoYOLO \cite{moosmann2023tinyissimoyolo},  
\item or reducing the required \gls{ram} needed for intermediate results while performing a network forward pass, for example, the different MCUNet versions \cite{mcunet, lin2021mcunetv2} \end{enumerate*}. 

Object detection models such as PP-PicoDet \cite{yu2021pp}, or NanoDet \cite{=nanodet} are promising networks but their smallest versions barely fit on \SI{1}{\mega\byte} of FLASH and exceed the commonly available \gls{ram} built-into \glspl{mcu}. Thus, these networks are benchmarked on Snapdragon class \glspl{soc} with power envelopes in the watt regime. 

Literature reports several networks for object detection on \glspl{mcu}, such as XiNet \cite{ancilotto2023xinet} and PhiNet\cite{paissan2022phinets} which are optimized for \gls{ram} consumption. However, these networks use layer operations such as attention mechanisms or squeeze and excitation blocks, that are not commonly implemented in commercially available accelerators built into \glspl{mcu} \cite{moosmann2023flexible}.
Most \gls{ml} accelerators, such as the \gls{rbe} accelerator implemented into the Marsellus \gls{soc} \cite{conti2023marsellus}, accelerate only a few, simple layer operations, for example, 3D convolution layers. Leveraging such accelerated operations, TinyissimoYOLO can achieve detection accuracy comparable to best-in-class models, while enabling real-time inference.

While the domain of object detection on edge devices, particularly within the constrained power envelope of \glspl{mcu}, has been explored to some extent, deploying small object detectors on such devices has remained relatively unexplored. Previous studies have primarily focused on optimizing existing neural network architectures for edge deployment or on reducing the computational complexity of object detection algorithms. However, the two-fold challenge of detecting small-sized objects on \glspl{mcu}, which necessitates both high detection accuracy and low computational overhead, has not been comprehensively addressed until now.

This paper makes several novel contributions that not only bridge the gap described in this section but also pushes the boundaries of what is currently achievable in terms of inference speed and detection accuracy on \glspl{mcu}:

\begin{itemize}
    \item \textbf{High-Speed Inference:} This paper extension demonstrates $>30$ \gls{fps} of inference execution using the \gls{fomo} network on low-power \glspl{mcu}. This significant achievement underscores our method's efficiency, making it viable for real-time applications where rapid object detection is critical.
    
    \item \textbf{State-of-the-Art Detection Accuracy:} With the deployment of the TinyissimoYOLO network on the GAP9 \gls{mcu}, this work reaches the current state-of-the-art mean object error count for small object detection. This is achieved while maintaining operational speeds of close to 3 \gls{fps}, a noteworthy accomplishment given the resource constraints of the target deployment platform.
    
    \item \textbf{Innovative Solution for Small Object Detection on \glspl{mcu}:} The methodology proposed in this paper extension addresses the untackled challenge of detecting small-sized objects on \glspl{mcu}. By optimizing both the network architecture and the inference pipeline, this paper extension presents a viable solution that meets the stringent requirements of high accuracy and low latency, which are paramount for edge-based applications.
\end{itemize}

% \todo[inline]{Juu \& Liam please do: we somewhere need to state how this related work influenced our decisions. So maybe after every paragraph quickly state one sentence for this...}

\section{Method}
\label{sec:method}
This section describes the different components that comprise the proposed adaptive tiling approach. An illustration of each step can be seen in \cref{fig:tiling_overview}.
\subsection{Detection Networks}

\subsubsection{\gls{fomo}}
\label{sec:fomo}
% \gls{fomo} \cite{fomo} is a lightweight object detection network that predicts object centers as opposed to bounding boxes. Solving this simplified problem results in a much smaller network than for example YOLO \cite{yolo}, which allows \gls{fomo} to run on \glspl{mcu}. \gls{fomo} uses the early layers of the MobileNetV2 \cite{mobilenetsv2} network as a feature extractor, resulting in an $n\times n$ feature grid, where the size of $n$ depends on how deep the feature extractor is. The implementation of \gls{fomo} used in this work has three downsampling stages resulting in an overall downsampling factor from input to output of 8. The feature grid is then fed into a small detection head that predicts object centers by classifying each of the $n\times n$ features and clustering predictions that belong to the same object. The original \gls{fomo} network, as implemented by \emph{Edge Impulse}, uses a crude clustering technique, whereby all predictions on neighboring grid cells are joined together. In the abstract image we show an overview of the \gls{fomo} architecture and an example of the output that this network generates.

\revdel{\gls{fomo} \cite{fomo} is a lightweight object detection network that predicts object centers instead of bounding boxes. It uses a feature extractor that generates a grid of $n\times n$ features, where the size of $n$ is determined by the input resolution. The features of each grid cell are classified into the different object classes.}

\revone{\gls{fomo} \cite{fomo} is a lightweight object detection network that predicts object centers as opposed to bounding boxes. \gls{fomo} uses the early layers of the MobileNetV2 \cite{mobilenetsv2} network as a feature extractor, resulting in an $n\times n$ feature grid, where the size of $n$ depends on how deep the feature extractor is. The implementation of \gls{fomo} used in this work has three downsampling stages resulting in an overall downsampling factor from input to output of 8. The feature grid is then fed into a small detection head that predicts object centers by classifying each of the $n\times n$ features and clustering predictions that belong to the same object. The original \gls{fomo} network, as implemented by \emph{Edge Impulse}, uses a crude clustering technique, whereby all predictions on neighboring grid cells are joined together.} %In the abstract image we show an overview of the \gls{fomo} architecture and an example of the output that this network generates.

This architecture has two key limitations. Firstly, it only makes one prediction for each of the $n\times n$ features, unlike other single-shot detectors, such as \gls{yolo} \cite{yolo}. This means that small objects which are localized in the same grid cell will count as one prediction. Secondly, it can not distinguish between nearby objects that are predicted in neighboring grid cells because the clustering step will incorrectly fuse them together.

Based on the application of carpark monitoring two modifications are made to the standard \gls{fomo} implementation. Because the ratio of objects to background can vary heavily from one image to another the weighted binary cross-entropy loss is replaced with a soft F1 loss introduced by Maiza et al. \cite{softf1}, which removes the need for manually tuning weights for the loss components of each object class. Furthermore, it can often be the case in carparks, that many of the cars are located in very close proximity. This makes it likely for predictions of neighboring cars to be located in neighboring grid cells of the \gls{fomo} output, which are then incorrectly joined by the \gls{fomo} fusion mechanism. In \cref{sec:fusion} we propose an alternative method for fusing predictions that minimizes accidental fusion of predictions belonging to separate objects.

\subsubsection{TinyissinoYOLO}
The work in \cite{liam_tiling} shows that the detection performance of \gls{fomo} with tiling stagnates for an input resolution of $192\times 192$. It is reasonable to assume that the small number of parameters of \gls{fomo} is responsible for this limitation thus, this work expands the results of \cite{liam_tiling} by investigating a more complex network.

TinyissimoYOLOv1.3 \cite{moosmann2023ultra} is an object detection network which
builds on TinyissimoYOLO \cite{moosmann2023tinyissimoyolo}, but uses the detection head of \gls{yolo}v3, giving it the ability to predict bounding boxes based on features extracted at multiple scales. Due to the simple network structure and its high utilization of 3D convolution layers, the network can be highly accelerated using \gls{soc} with \gls{ml} accelerators as shown by the successor publication of TinyissimoYOLO \cite{moosmann2023flexible}. The ability of this network to predict bounding boxes allows for more sophisticated algorithms to be used to fuse predictions from neighboring tiles, as will be outlined in \cref{sec:fusion}.
%TinyissimoYOLO \cite{moosmann2023tinyissimoyolo} is an object detection network utilizing the \gls{yolo}v1 backbone and detection head with reduced size. Due to the simple network structure and its high utilization of 3D convolution layers, the network can be greatly accelerated using \gls{soc} with \gls{ml} accelerators as shown by the successor publication of TinyissimoYOLO \cite{moosmann2023flexible}. In this publication, we utilize the TinyissimoYOLOv1.3 network \cite{moosmann2023ultra} which uses the backbone of \gls{yolo}v1 and the detection head of \gls{yolo}v3 in combination with the previously proposed tiling method. It has been shown \cite{moosmann2023ultra}, that this network has still relatively high object detection accuracy with low latency of \SI{16.2}{\milli\second} on a GAP9 \gls{mcu}. In contrast to the \gls{fomo} network, which network's size and thus knowledge capacity seems to stagnate with input images of 192$\times$192 pixel resolution while the TinyissimoYOLOv1.3 network can handle images with higher resolution of up to 256$\times$256 pixels. This increases the relative object size of the used carpark monitoring application, making it more confident on small sized object detection predictions. 
%Further, the prediction of bounding boxes is very attractive for a tiling approach, making the post-tiling-prediction algorithms applied for fusing all the predictions of the tiled images more sophisticated.
% \todo[inline]{julian: Add short paragraph about TinyissimoYolo?}

\subsection{Adaptive Tiling}
\begin{figure}
\centering
\includegraphics[width=0.45\textwidth]{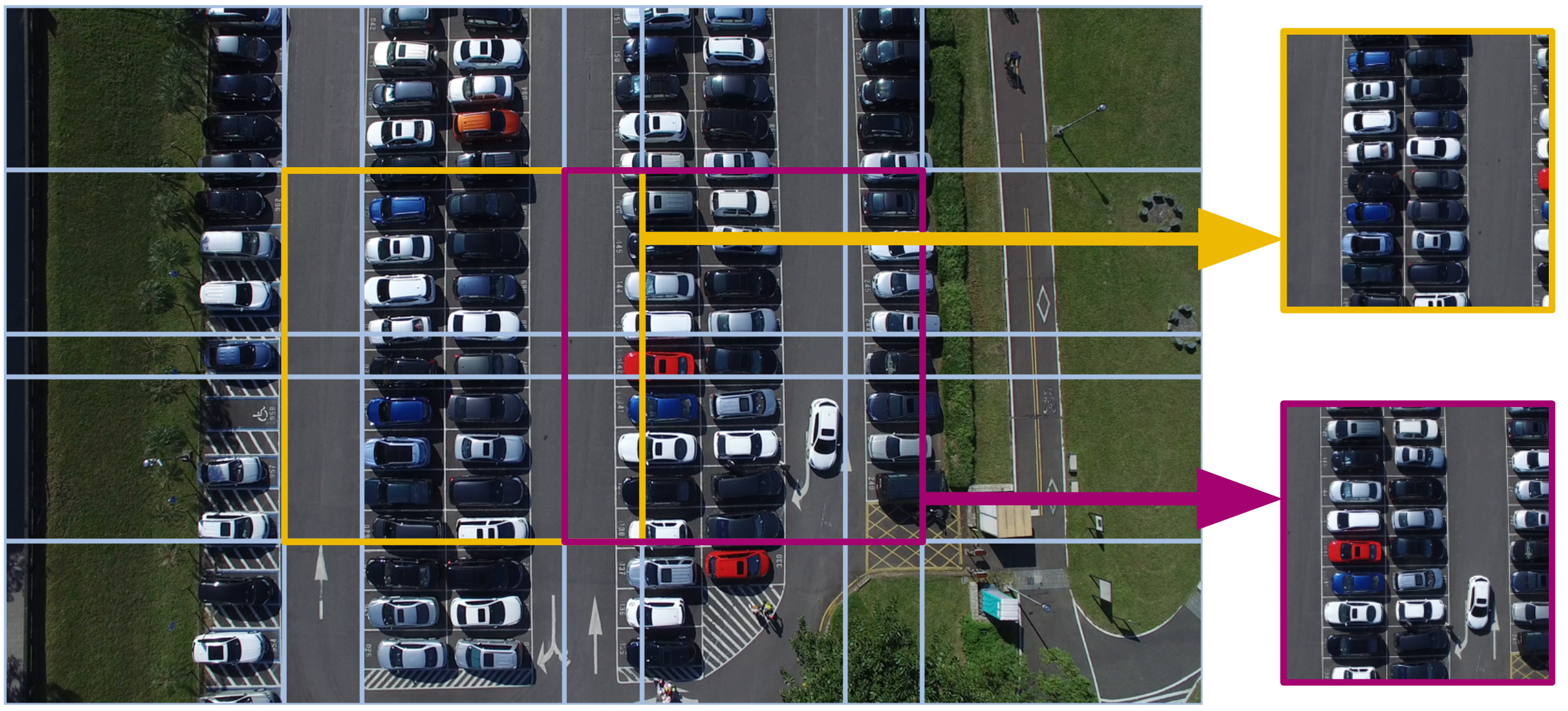}
\caption{Example image split into 12 overlapping tiles. Tiles are extracted from the image and downsampled to the network input resolution. This example demonstrates how objects, that are cut off by the tile border (see yellow tile) %as is the case with the row of cars on the right in the red tile, 
are fully in view in the neighboring tile thanks to the tiles being overlapping.}
\label{fig:tiling-sheep}
\vspace{-10pt}
\end{figure}

To address the general issue of lacking pixel information for small objects as well as the object size limitation stemming from the \gls{fomo} architecture, we propose an adaptive tiling approach to split images into smaller sub-images. This approach has multiple advantages. Firstly, it increases the relative size of the objects thus increasing the signal-to-noise ratio assuming that we keep the input resolution to the network constant. Secondly, increasing the size of the objects reduces the probability of multiple objects being located in the same grid cell, which is very important for anchor-based object detectors like \gls{fomo} and TinyissimoYOLO. %Furthermore, with the relative increase in size of the objects one can counteract the bias in most datasets towards larger objects.
We follow the work of Akyon et al. \cite{sahi} and split the images into overlapping tiles, ensuring that objects which are only partially visible in one tile are fully visible in the neighboring tile. \cref{fig:tiling-sheep} shows how the proposed method divides an image into smaller tiles for individual processing. %This example demonstrates how the tiling increases the relative size of the objects and the highlighted tiles exemplify how the overlapping avoids objects being split by the tileborder in both tiles.

The drawback of splitting images into tiles for individual processing is the resulting increase in processing time and the smaller the objects are, the more tiles need to be used to get good detection results. In this work a target object size is defined, which is quantified using the \gls{nba}, to adjust the number of tiles. The \gls{nba} is calculated as the area, in pixels, of a bounding box, \textit{normalized} by the total area of the image in pixels. For images taken close to the objects, only a few tiles are necessary to achieve the target object size, and vice versa for images taken far away from the objects. This minimizes the number of tiles needing to be processed while ensuring the best possible detection performance.

\revdel{For a specific image, and a given target object size, a square tile is defined such that}

\revone{To ensure that the objects in the tiled image match the target object size we define a square tile for each image with}

\begin{equation}
    tile_{wh} = \sqrt{\frac{a_{image}\cdot i_{NBA}}{t_{NBA}}} ,
\end{equation}
\label{eq:tile_wh}

where \revone{$tile_{wh}$ indicates the tile width/height,} $i_{NBA}$ is the average object size, $a_{image}$ is the image area in pixels and $t_{NBA}$ represents the target object size. Finally, the numbers of tiles in x and y dimensions are chosen such that the tile overlap is at least 1.5 times the average object width/height. 

%In \cref{fig:carpk_tiling_nba_hist_0_006} we show the effect that our tiling approach has on the object size distribution of the dataset. The top plot shows the size distribution for the original dataset and in the bottom plot, we show the reshaped distribution after applying our tiling with a target object size of 0.6\%.

During training, the bounding box sizes of the labeled objects are used to calculate the optimal tile dimensions. In real-world applications, one could use height measurements from GPS for example, to estimate the object size.

% \begin{figure}
% \centering
% \includegraphics[width=0.5\textwidth]{figures/carpk_tiling_nba_hist.png}
% \caption[Tiling Effect on Bbox area]{The left histogram shows the distribution of \gls{nba} values for the train and test sets without any tiling. On the right, we show the resulting distribution of object sizes when applying the adaptive tiling step with a target \gls{nba} value of 0.006. }
% \label{fig:carpk_tiling_nba_hist}
% \end{figure}

% \begin{figure}
% \centering
% \includegraphics[width=0.45\textwidth]{figures/tiling_hist_0_006_vert.png}
% \caption[Tiling Effect on Bbox area]{The top histogram shows the distribution of \gls{nba} values for the train and test sets without any tiling. At the bottom, we show the resulting distribution of object sizes when applying the adaptive tiling step with a target \gls{nba} value of 0.006. }
% \label{fig:carpk_tiling_nba_hist_0_006}
% \end{figure}

\subsection{Fusing Tiled Predictions}
\label{sec:fusion}
Using overlapping tiles results in image areas that are seen multiple times by the network, hence the network is likely to make multiple predictions for the same objects. To ensure an accurate object count, predictions of the same object from multiple tiles must be fused correctly. This is achieved by comparing the intersections of predictions from all tiles, however, there are slight implementation differences depending on the network architecture.

% As explained in \cref{sec:fomo}, the implementation of \gls{fomo} used in this work makes centroid predictions on an $n\times n$ grid where $n$ is dependent on the network input resolution. For an input size of $192\times 192$ the resulting output shape is $24\times 24$, where the downsampling factor of $8$ stems from the three downsampling stages that can be seen in in the abstract image. This results in $576$ centroid predictions per tile. For each of these $576$ predictions, we create a "pseudo" bounding box that is the size of the corresponding grid cell. Assuming an input size of $192\times 192$ and $24\times 24$ predictions the resulting bounding boxes are of shape $8\times 8$ pixels. Finally, we fuse all bounding boxes with \gls{iou} greater than a specific threshold.% which is set to $0.075$.

%As explained in \cref{sec:fomo}, the implementation of \gls{fomo} used in this work makes centroid predictions on an $n\times n$.
For each of \revone{the}\revdel{these} $n\times n$ predictions made by the \gls{fomo} network, a "pseudo" bounding box is created that is the size of the corresponding grid cell. These grid-cell-sized bounding box predictions are fused between neighboring tiles when their \gls{iou} is greater than a specific threshold.

\begin{figure}
\centering
\includegraphics[width=0.45\textwidth]{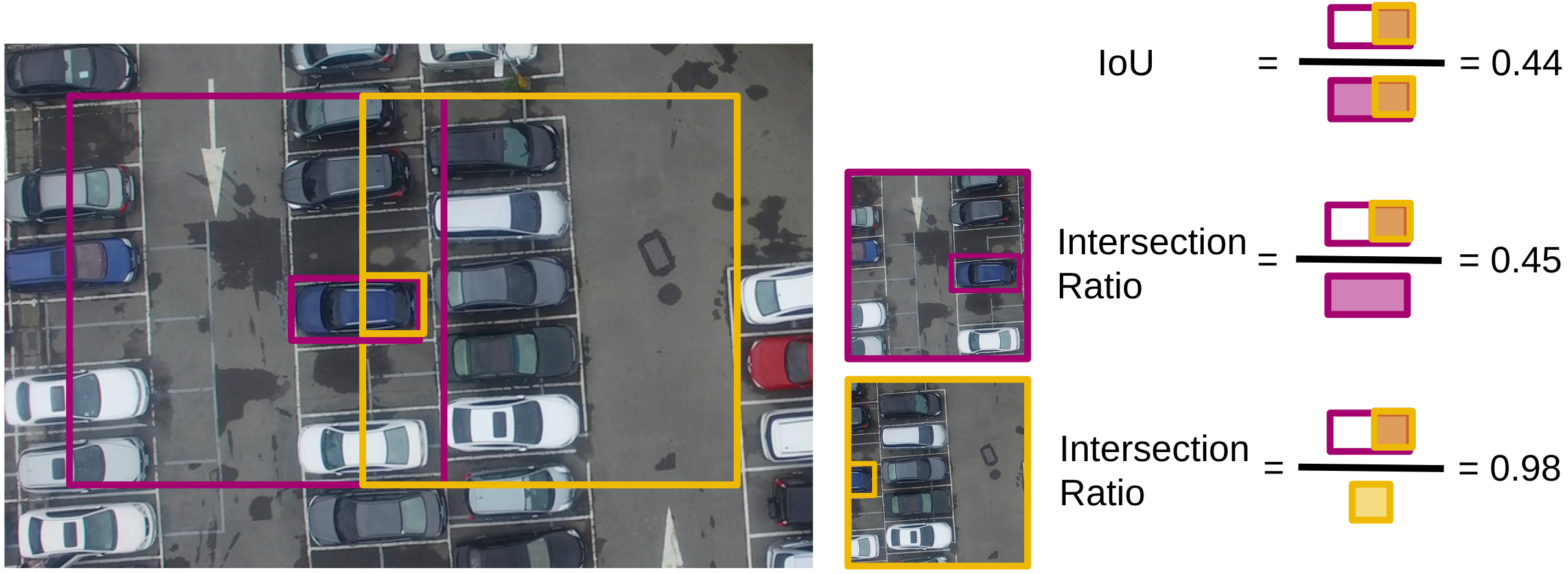}
\caption{The figure illustrates the matching criterion for predictions made by TinyissimoYOLO. The blue car is only partially visible inside the yellow tile thus, the network can only predict a bounding box that only covers part of the car. Since neighboring tiles are overlapping, the same car is fully visible in the purple tile resulting in a bounding box prediction that covers the whole car. The right side shows the calculation of \gls{iou} and intersection ratios for both tiles.}
\label{fig:yolo_fusion}
\vspace{-10pt}
\end{figure}

TinyissimoYOLO also makes predictions on a grid, however, rather than predicting centroid locations, TinyissimoYOLO predicts bounding boxes for each of the $n\times n$ anchor points. One could still use the \gls{iou} metric to determine correspondences between overlapping predictions, however, when objects are only partially visible in one tile but fully visible in a different tile, the bounding boxes for the same object will be of very different sizes. An example of such a case can be seen in \cref{fig:yolo_fusion}. This example shows that even if a bounding box from one tile is fully overlapping with a second bounding box in a different tile, the \gls{iou} can be quite small. This makes it impossible to find a threshold for which nearby objects are not accidentally fused while ensuring a correct fusion of predictions belonging to the same object.

To \revone{address}\revdel{avoid} this problem we use a slightly different matching approach. Instead of a two-way matching based on the \gls{iou}, we use a one-way matching based on \revone{what we call the \textit{intersection ratio}. This ratio is determined for each bounding box by dividing the intersection area with another box by its own area. The desirable property of this metric is that} \revone{bounding boxes which are almost entirely enclosed by another box, will have an intersection ratio close to 1.}\revdel{the ratio of the intersection and the area of the bounding box itself.} If the intersection ratio is larger than a threshold for at least one of the bounding boxes the pair is considered a match. An example of this calculation can be seen in \cref{fig:yolo_fusion}, where the yellow tile can only partially see the object and thus predicts a smaller bounding box compared to the purple tile, resulting in an \gls{iou} of less than $50\%$. On the other hand, the intersection ratio of the prediction in the yellow tile is $98\%$ and thus the two predictions can be considered a match with high confidence when using our one-way matching.

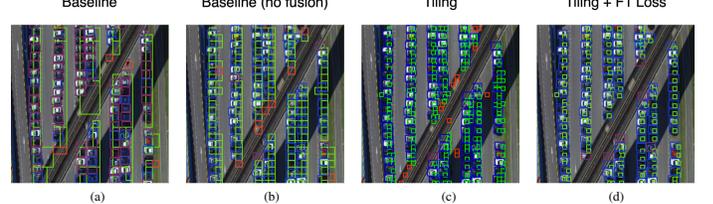
\begin{figure}
\centering
\begin{overpic}[width=0.5\textwidth]{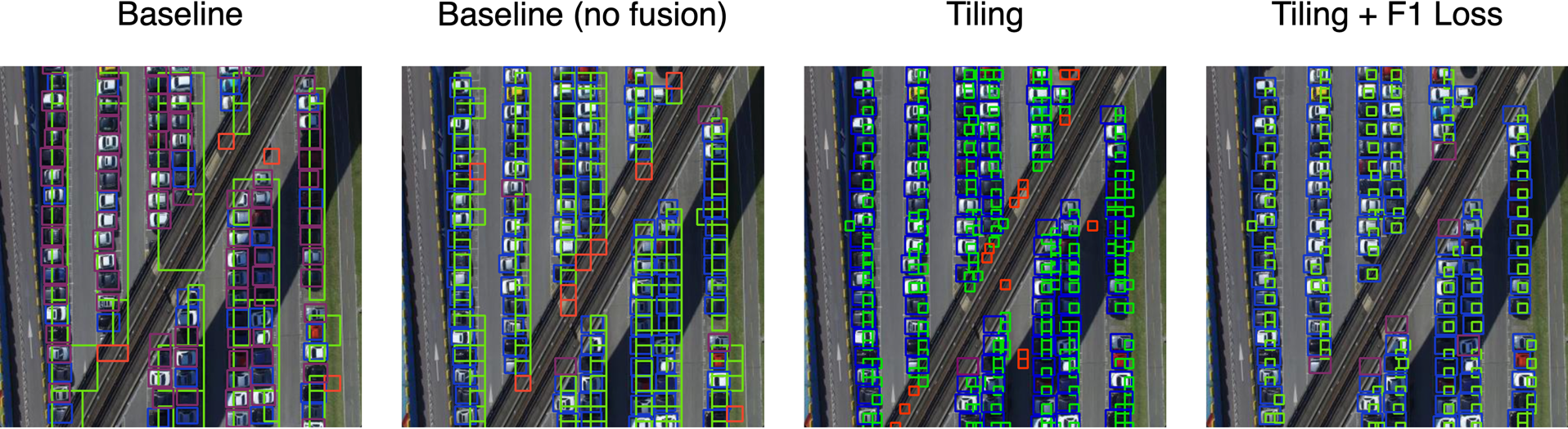}
        \put(11.5,-2.5){\textcolor{black}{{\tiny (a)}}}
        \put(37,-2.5){\textcolor{black}{{\tiny (b)}}}
        \put(63,-2.5){\textcolor{black}{{\tiny (c)}}}
        \put(87.5,-2.5){\textcolor{black}{{\tiny (d)}}}
\end{overpic}
\caption{The green rectangles indicate true positive predictions, the blue rectangles are ground truth object bounding boxes that were correctly predicted, the red rectangles show false positive predictions and lastly the purple rectangles are ground truth bounding boxes for cars that were not identified. In subfigure a) and b) we show the prediction output of the baseline model which is the default \gls{fomo} network with and without the \gls{fomo} fusion method, respectively. Subfigures c) and d) depict the output of \gls{fomo} using our tiling method. In d) one may observe that training with a soft F1 loss significantly reduces the number of false positives. This graphic is reused from Boyle et al. \cite{liam_tiling}.}
\label{fig:ablation}
\vspace{-10pt}
\end{figure}

\section{Results}
We evaluate the detection performance of the standard \gls{fomo} implementation \cite{fomo}, our implementation of \gls{fomo}, and TinyissimoYOLO \cite{moosmann2023tinyissimoyolo}, both with the proposed adaptive tiling method on the CARPK \cite{carpk} dataset. This dataset was collected by a drone flying at roughly 40 meters altitude over four different parking lots. It consists of close to 1,500 images containing more than 90,000 cars.

\subsection{Experimental Settings}
\revone{As well as comparing the proposed method to other work on the CARPK \cite{carpk} dataset we also evaluate the performance gains, in terms of small object detection, that arise from different configurations of our tiling method. More specifically, we train the \gls{fomo} models with three different input resolutions as well as four different values of target \gls{nba}, while TinyissimoYOLOv1.3 is trained with one input resolution, and four larger target \glspl{nba} values. In this paper, the GAP9 \gls{mcu}---equipped with a cluster of 9 \emph{RISC-V} cores, one fabric controller, 1.6 MB of on-chip \gls{ram}, and 2MB of on-chip non-volatile memory---is used for the deployment and evaluation. The investigated models are quantized to Int8 precision by using the NNTooL\footnote{\url{https://github.com/GreenWaves-Technologies/gap_sdk/blob/master/tools/nntool}}. The NNTool is used to optimize the computational graph, quantizing and validating the topology. The Autotiler\footnote{\url{https://github.com/GreenWaves-Technologies/gap_sdk/tree/master/tools/autotiler_v3}} is used for automatic c-code generation based on the topology-optimized and quantized graph. The code generated by the Autotiler is then used for executing the quantized networks on the device. The latency and energy measurements are conducted by deploying the networks on the GAP9 evaluation kit and utilizing the Power Profiler Kit 2 from \emph{Nordic Semiconductor}.  Deploying the models onto GAP9 allows us to observe how---not only the detection performance but also---the memory consumption and the latency is affected by different target object sizes of the proposed tiling method as well as the varying input resolutions to the network.}

\revone{The deployed networks are evaluated and compared by number of parameters, latency, inference efficiency---which describes how well the computational workload is parallelized on the device--- and energy per inference as suggested by Giordano et al. \cite{giordano2022survey}.}

\subsection{FOMO}
 In \cref{tab:count_error} we compare the object count error, the F1 score, and the model size of \gls{fomo} using our tiling method against the standard \gls{fomo} implementation as well as other published results on the CARPK dataset. Our approach achieves a reduction of $76\%$ in the object count error and a $225\%$ increase in F1 score compared to the original \gls{fomo} architecture. The \gls{mae} for \gls{yolo} is reported from \cite{carpk} and the value for \gls{yolo}v4 is reported from \cite{sf_ssd}. For both \gls{yolo} and \gls{yolo}v4 we estimate the number of parameters based on the implementations in \cite{yolo} and \cite{yolov4}, respectively. Compared to \gls{yolo} \cite{yolo} our implementation of \gls{fomo} with tiling reduces the \gls{mae} by $74\%$, while \gls{yolo}v4 \cite{yolov4} achieves a $85\%$ reduction. However, both \gls{yolo} and \gls{yolo}v4 are not designed for \glspl{mcu} and do not fit the memory requirements of most low-power embedded devices. This is showcased clearly in \cref{tab:count_error} when comparing the number of model parameters.

\begin{table}
\centering
\caption[Comparison of Count Error on CARPK Dataset]{Count error (MAE) comparison for the implementation of \gls{fomo} with tiling, indicated by the *, TinyissimoYOLO with tiling as well as the standard \gls{fomo} implementation with previous work on the CARPK dataset.}
\begin{tabular}{ll|l|l|l}
\toprule
\toprule
& Method       & MAE           & F1   & Parameters\\
\midrule
\midrule
&\gls{fomo}         & 54,32         & 0.28 & \textbf{19K}\\
&\gls{yolo}         & 48,89         & - & 119M\\
&\gls{yolo}v4       & 7,16         & - & 35M\\
\midrule
\multirow{2}{*}{\rotatebox[origin=c]{90}{\textbf{Ours}}} & FOMO* + Tiling         & 12,9          & \textbf{0.91} & \textbf{19K}\\
& TinyissimoYolo + Tiling         & \textbf{6.2}          & \textbf{0.91} & 403K\\
\bottomrule
\bottomrule
\end{tabular}

\label{tab:count_error}
\vspace{-10pt}
\end{table}

\begin{table}
\centering
\caption[Ablation Study]{Ablation study conducted on the CARPK dataset. BL indicates the baseline with standard \gls{fomo} implementation from \cite{fomo}, nF indicates using no fusion, T stands for tiling and F1 Loss indicates the usage of F1 loss during training.}
\begin{tabular}{l|l|l|l|l|l|l|l}
\toprule
\toprule
BL    & nF            & T      & F1 Loss      & F1             & Pr            & Re            & MAE            \\
\midrule
\midrule
\checkmark  &                      &             &             & 0,28          & 0,87          & 0,17          & 54,32          \\ 
% \hline
\checkmark  & \checkmark           &             &             & \textbf{0,95} & 0,96          & \textbf{0,95} & 34,76          \\ 
% \hline
\checkmark  & \checkmark           & \checkmark  &             & 0,94          & 0,95          & 0,94          & 44,17          \\ 
% \hline
\checkmark  & \checkmark           & \checkmark  & \checkmark  & 0,91          & \textbf{0,99} & 0,85          & \textbf{12,8}  \\
% \hline
% \checkmark  & \checkmark           & \checkmark  & \checkmark & \checkmark    & 0,89          & \textbf{0,99} & 0,82          & 13,4  \\
\bottomrule
\bottomrule
\end{tabular}
\label{tab:ablation}
\end{table}

\begin{figure*}
\centering
\begin{overpic}[width=\textwidth]{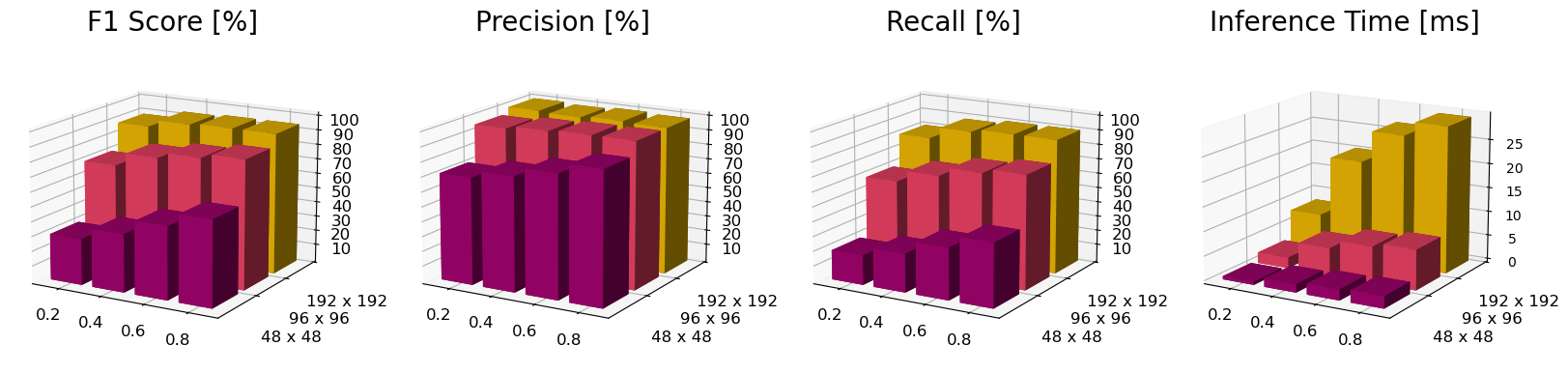}
        \put(5,0){\textcolor{black}{{\tiny Target NBA $[\%]$}}}
        \put(18,0){\textcolor{black}{{\tiny Resolution}}}
        \put(30,0){\textcolor{black}{{\tiny Target NBA $[\%]$}}}
        \put(43,0){\textcolor{black}{{\tiny Resolution}}}
        \put(55,0){\textcolor{black}{{\tiny Target NBA $[\%]$}}}
        \put(68,0){\textcolor{black}{{\tiny Resolution}}}
        \put(80,0){\textcolor{black}{{\tiny Target NBA $[\%]$}}}
        \put(93,0){\textcolor{black}{{\tiny Resolution}}}
\end{overpic}
\caption{This figure shows the resulting F1 metrics as well as the latency for different configurations of our implementation of \gls{fomo}.}
\label{fig:latency}
% \vspace{-10pt}
\end{figure*}
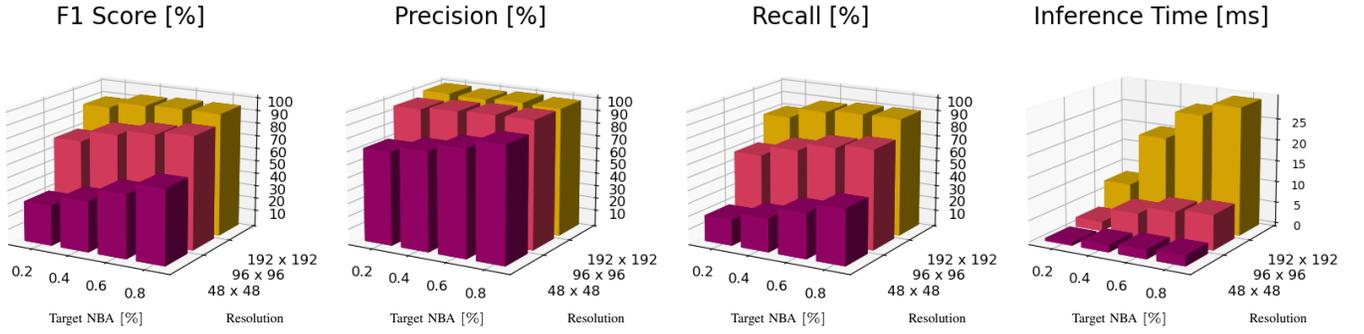

\subsubsection{On Device Performance}
In \cref{fig:latency} the F1 metrics and the on-device latency are shown for different configurations of our method. 
From the data in \cref{fig:latency} one can observe that both, increasing the target \gls{nba}, i.e. using more tiles, as well as using a higher input resolution results in significantly improved F1 metrics. The strong improvement in detection performance suggests that the main obstacle in small object detection is indeed the relative lack of information that is available for small objects. These results suggest, when increasing the input resolution is not an option\textemdash due to memory restrictions\textemdash the proposed tiling approach is a valid alternative to boost the signal-to-noise ratio of small objects and with that, the detection performance. What we can also see in these results is the additional latency that is incurred by the tiling method. For example, reaching the target \gls{nba} of $0.8\%$ requires 3.66 times more tiles than the target \gls{nba} of $0.2\%$,
which results in the latency increasing from \SI{8.0}{\milli\second} to \SI{30.0}{\milli\second} when processing a full image with a network input resolution of $192\times 192$.

%which results in a total latency increase of 21.95 milliseconds to process an entire image, when using an input resolution of $192\times 192$.

\subsection{TinyissimoYOLO}
As can be seen in the previous work of this extension and again in \cref{fig:latency} the detection performance of \gls{fomo} stagnates at a target \gls{nba} of $0.8\%$ and an input resolution of $192\times 192$. This stagnation can be attributed to the extremely low number of parameters in the \gls{fomo} network, which limits its prediction ability. For this reason, experiments are conducted using the TinyissimoYOLOv1.3 \cite{moosmann2023ultra} network in combination with our tiling approach. This network is considerably larger, which enables it to make bounding box predictions as opposed to predicting centroids only. 

In \cref{tab:tiny_yolo} the results of the tiling method in combination with TinyissimoYOLOv1.3 for different target object sizes is reported. Using a target \gls{nba} of $0.8\%$, which is the value at which the performance of \gls{fomo} stagnates a precision and recall of $93\%$ and $46\%$, respectively are achieved. The low recall results in an object count error of $53.6$, which is significantly worse than what \gls{fomo} achieves. However, unlike \gls{fomo}, increasing the target object size beyond $0.8\%$ brings significant improvements. With a target \gls{nba} of $4\%$ the F1 score of the best performing \gls{fomo} model is matched and an object count error of $6.2$ is achieved. This represents a reduction of $52\%$ compared to our best performing \gls{fomo} network and it even outperforms large-scale models like YOLOv4 \cite{yolov4}, as seen in \cref{tab:count_error}. %The last row of \cref{tab:tiny_yolo} shows the detection results for tiling with a target \gls{nba} of $6\%$, where a slight decrease in all metrics but recall can be seen. At a target \gls{nba} of $6\%$ images are being split into over 30 tiles on average, which leads to a large number of cars being split by the many tile borders. Only partially seeing so many of the objects ends up decreasing the overall detection performance, thus we can say that this is the limit of tiling that makes sense to use.

The final row of \cref{tab:tiny_yolo} illustrates the detection results obtained when employing a target \gls{nba} of $6\%$, revealing a slight decrease across all metrics except for recall. With a target \gls{nba} of $6\%$, images are partitioned into almost 30 tiles on average, resulting in many cars being intersected by multiple tile boundaries, reducing the detection performance. Hence, we assert that this represents the practical limit of tiling where the benefit of increasing the object size is outweighed by the many objects being intersected by tile boundaries.

\begin{table}
\centering
\caption[Performance of TinyissimoYOLO with tiling]{Performance comparison of TinyissimoYOLO for different target object sizes (NBA). For each target \gls{nba} we report the average number of tiles an image is split into, the resulting \gls{fps} for running inference on all tiles that comprise one image, the object count error (MAE), the F1 score (F1), precision (Pr) and recall (Re).% (Target \gls{nba}. All networks use an image size of $256\times 256$.
}
\resizebox{\columnwidth}{!}{
\begin{tabular}{l|l|l|l|l|l|l}
\toprule
\toprule
Target NBA [\%] & Avg \#Tiles & Count MAE    & F1            & Pr            & Re           & \gls{fps} \\
\midrule
\midrule
0.8             &4.1         & 53.6         & 0.61          & \textbf{0.93} & 0.46         & 15.1\\
2.0             &10.1        & 21.2         & 0.82          & 0.92          & 0.74         & 6.1\\
4.0             &21.6        & \textbf{6.2} & \textbf{0.91} & \textbf{0.93} & \textbf{0.89}& 2.9\\
6.0             &27.8        & 6.6          & 0.90          & 0.90          & \textbf{0.89}& 2.2\\
\bottomrule
\bottomrule
\end{tabular}
\label{tab:tiny_yolo}
}
\end{table}

% \begin{figure}
% \centering
% \includegraphics[width=0.5\textwidth]{figures/tinyYOLO_prediction.png}
% \caption{The green rectangles indicate true positive predictions, the blue rectangles are ground truth object bounding boxes that were missed, i.e. false negatives, and the red rectangles show false positive predictions.}
% \label{fig:label_inconsistencies}
% % \vspace{-10pt}
% \end{figure}
% \todo[inline]{Here we can describe the results we get from using TinyissimoYolo. The bare minimum would be showing an example prediction image and stating the achieved metrics. Ideally we would also get results from a quantized model such that we can compare with the quantized FOMO. Here we could also show some of the inconsistencies in the CARPK labelling.}

\subsubsection{On Device Performance}
\begin{figure}[h!]
\centering
\includegraphics[width=0.49\textwidth]{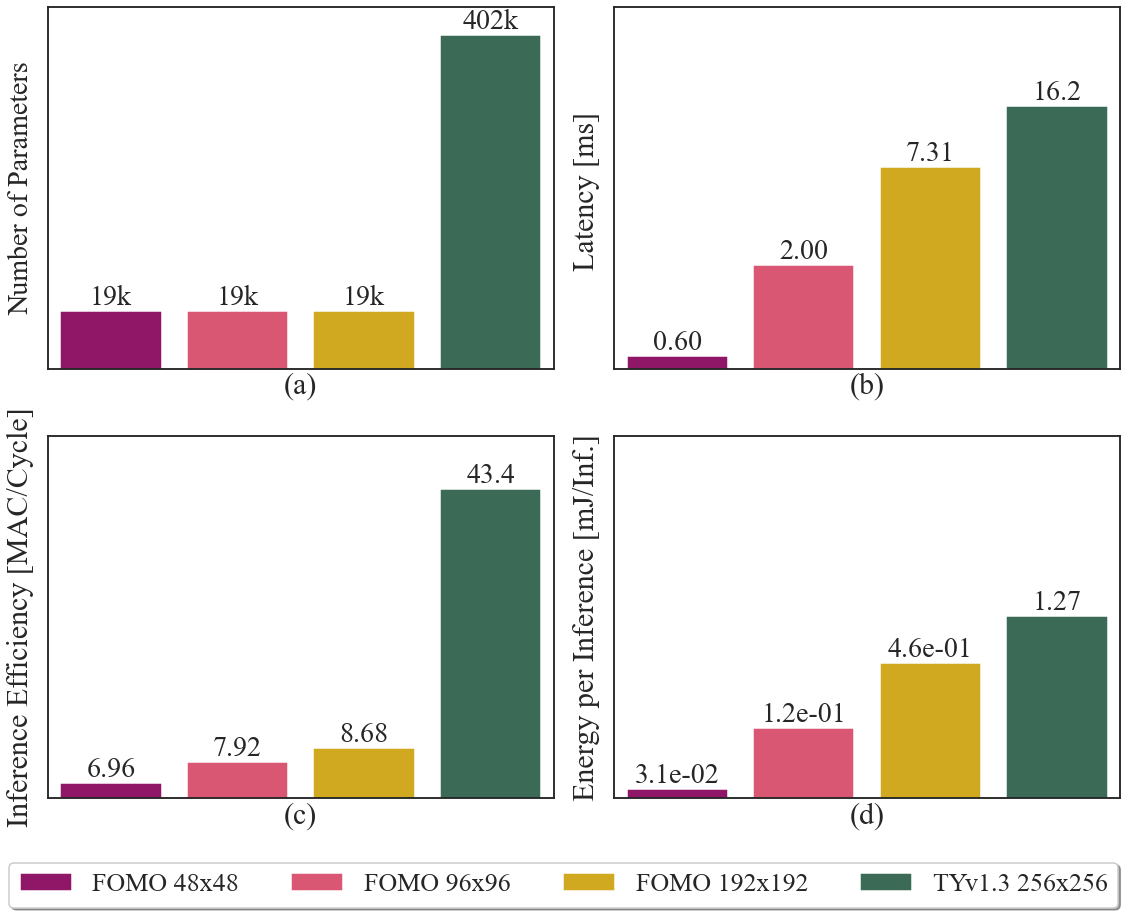}
\caption{The three \gls{fomo} networks and TinyissimoYOLO deployed on GAP9's NE16 accelerator (@\SI{370}{\mega\hertz}, 1.8V) with indicated input image resolution and compared in terms of a) parameter count, b) latency, c) inference efficiency and d) energy consumption per inference.}
\label{fig:gap9_benchmark}
% \vspace{-10pt}
\end{figure}

By deploying the Int8 quantized \gls{fomo} networks and TinyissimoYOLO on GAP9's NE16 accelerator, real-time inference is achieved. \Cref{fig:gap9_benchmark} compares the deployed networks in terms of the metrics described in \cref{sec:method}. In \cref{fig:gap9_benchmark}a) it can be observed that TinyissimoYOLO is roughly 21$\times$ larger in size compared to the implemented \gls{fomo} networks. However, GAP9 parallelizes the workload of TinyissimoYOLOv1.3 5$\times$ better than the workload of FOMO, see \cref{fig:gap9_benchmark}c) resulting in a 2.2$\times$ longer execution time for a network that is 21$\times$ bigger. This discrepancy can be explained by the NE16's capability of accelerating especially 3D convolutions, which are the core layers of the TinyissimoYOLOv1.3 network. As a consequence of the accelerator's energy efficiency the energy consumption of TinyissimoYOLOv1.3, running on GAP9, increases by only 2.7$\times$ compared to \gls{fomo}, as seen in \cref{fig:gap9_benchmark}d).

\subsection{Ablation}

% As described in \cref{sec:fomo}, multiple changes are made to the standard \gls{fomo} implementation besides the adaptive tiling step. Here we give insights on the effect that each of these changes has on the detection performance of \gls{fomo}.
\revone{Besides adding the adaptive tiling step\revone{,} two changes were made to the standard \gls{fomo} implementation as described in \cref{sec:fomo}. The changes include (1) removing the fusion step and (2) incorporating a soft F1 loss during training. In \cref{tab:ablation} we showcase the effect these changes have on the object count error as well as the F1 metrics. The baseline for this comparison is the standard \gls{fomo} implementation and we use an input resolution of $192\times 192$ for all networks.}

\revone{\cref{fig:ablation} compares the output of all networks for the same input image. \cref{fig:ablation}a) highlights the negative impact that the \gls{fomo} fusion method has.} \revone{The cars in this scene are parked tightly together such that even the ground truth bounding boxes of neighboring cars touch each other. This means that most of the cars lie in neighboring grid cells of the \gls{fomo} feature grid, which leads \gls{fomo} to incorrectly fuse many predictions into large bounding boxes spanning multiple cars. As can be seen in \cref{tab:ablation} removing this fusion step improves the recall as well as the object count error by large margins. These results are confirmed visually in \cref{fig:ablation}b), where it can be seen that removing this fusion step drastically reduces the number of false negatives.}

\revone{The \cref{fig:ablation}c) exemplifies how the proposed tiling method both increases the number and reduces the size of the \gls{fomo} prediction grid cells. This process reduces the likelihood of multiple cars being covered by the same grid cell while at the same time increasing the likelihood of one car being covered by multiple grid cells. The proposed fusion method described in \cref{sec:fomo} is capable of fusing overlapping predictions but it is not able to fuse predictions belonging to the same car when they don't overlap, leading to an inflated object count. This explains why the object count in \cref{tab:ablation} gets worse, while the F1 metrics remain at a similar level.}

Finally, \cref{fig:ablation}d) shows the effect that training with a soft F1 loss has. Using this loss, the network is encouraged to make only one prediction per object, even when it is covered by multiple grid cells. This seemingly "implicit" non-maximum suppression greatly reduces the duplicate predictions per object and results in a significant reduction in the object count error such that, compared to the standard \gls{fomo} implementation, we achieve an increase in F1 score of $225\%$ and a reduction of the object count error by $76\%$.

%However, the latency of TinyissimoYOLO, reported in \cref{fig:gap9_benchmark}b), is only about 2.2 $\times$ slower than the slowest \gls{fomo} network. 

%While \cref{fig:gap9_benchmark}a) denotes the network's sheer size difference, the latency \cref{fig:gap9_benchmark}b) is not scaling linearly with the size. This is due to the NE16's capability of accelerating especially 3D convolutions, which are the core layers of the TinyissimoYOLOv1.3 network. GAP9 parallelizes the workload of TinyissimoYOLOv1.3 5$\times$ better than the workload of FOMO, see \cref{fig:gap9_benchmark}c) resulting in a 2.2$\times$ longer execution time for a network 21$\times$ bigger. Interestingly, the energy efficiency due to NE16's workload parallelization results in an energy consumption increase of only 2.7$\times$ comparing \gls{fomo} with $192\times192$ image resolution to the TinyissimoYOLO network, see \cref{fig:gap9_benchmark}d).

% \todo[inline]{Julian: This section needs to be expanded based on the latency/power measurements we get from GAP9 using \gls{fomo} and TinyissimoYolo}

\section{Conclusion}
\textcolor{black}{This paper extends the adaptive tiling method introduced in our previous work \cite{liam_tiling} by applying it to the latest TinyissimoYOLOv1.3 network. The bounding box predictions made by TinyissimoYOLOv1.3 can be utilized by the adaptive tiling method to smartly fuse predictions from overlapping tiles. The novel tile fusion algorithm results in a reduction of the mean object count error from the previously achieved 12.9 MAE, using our implementation of \gls{fomo} \cite{liam_tiling}, down to 6.2 MAE, reaching state-of-the-art detection accuracy on the CARPK dataset \cite{carpk}.} \textcolor{black}{This work demonstrates edge deployment on the\revdel{RISC-V} \revone{\emph{RISC-V}}-based GAP9 \gls{mcu} with built-in \gls{ml} accelerator from\revdel{Greenwaves' Technology} \revone{\emph{GreenWaves Technologies}}. The best-performing implementation of \gls{fomo} can run inference on a single tile in \SI{7.31}{\milli\second} allowing it to predict object locations on full images at 33.4 \gls{fps}, while TinyissimoYOLOv1.3 requires \SI{16.2}{\milli\second} inference time per tile, resulting in a prediction rate of 2.8 \gls{fps} on full images.} \textcolor{black}{We demonstrate how the proposed tiling method can be used for high detection performance, rivaling state-of-the-art large-scale networks on the CARPK dataset \cite{carpk}, while being able to fulfill the stringent low memory and low power requirements of \glspl{mcu}.
These results represent a significant step towards small object detection on \glspl{mcu}, running in real-time.}

\section*{Acknowledgments}
The authors would like to thank armasuisse Science \& Technology and RUAG Ltd. for funding this research.

\bibliographystyle{IEEEtranDOI}
\bibliography{main.bib}

\begin{IEEEbiography}[{\includegraphics[width=1in,height=1.25in,clip,keepaspectratio]{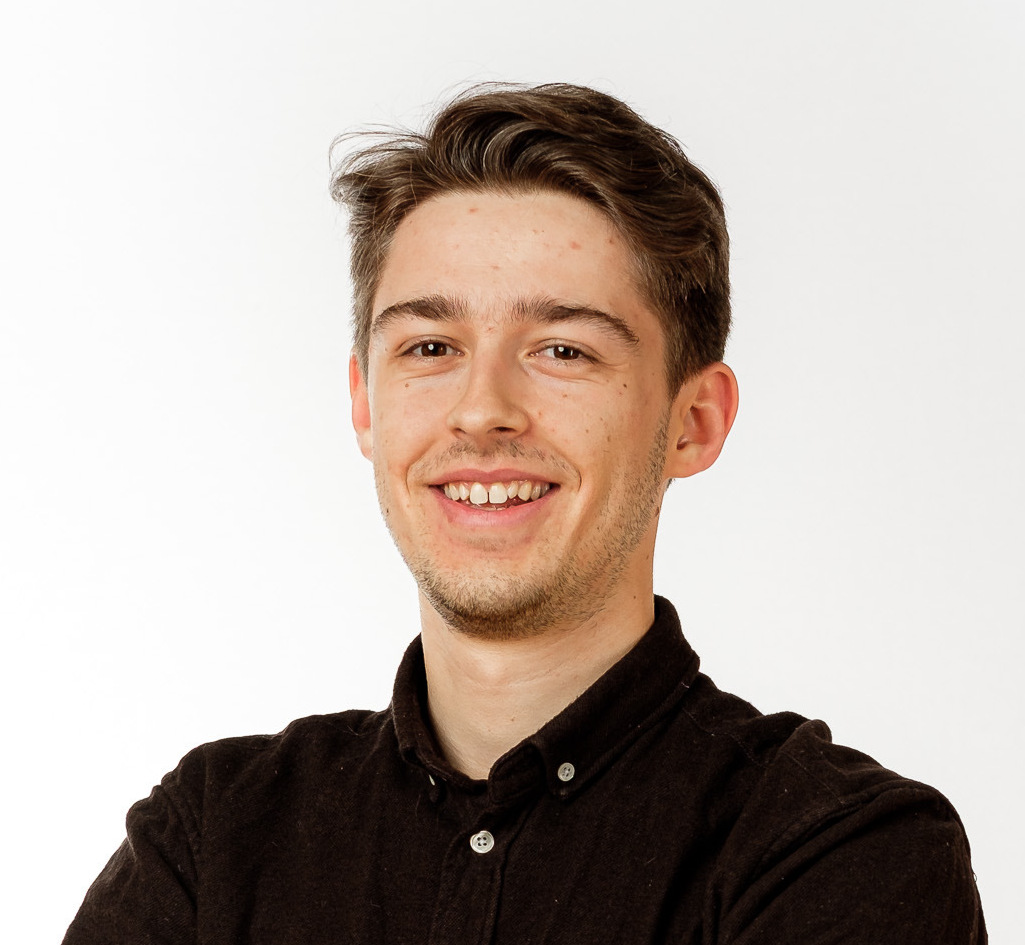}}]{Liam Boyle}
received his BSc and MSc degrees in mechanical engineering from of ETH Zurich, Zürich, Switzerland, in 2020 and 2024, respectively. Currently, he is a research assistant at the Center for Project-Based Learning at ETH Zurich. His research regards computer vision and sensor fusion in the domain of robotics.
\end{IEEEbiography}

\begin{IEEEbiography}
[{\includegraphics[width=1in,height=1.25in,clip,keepaspectratio]{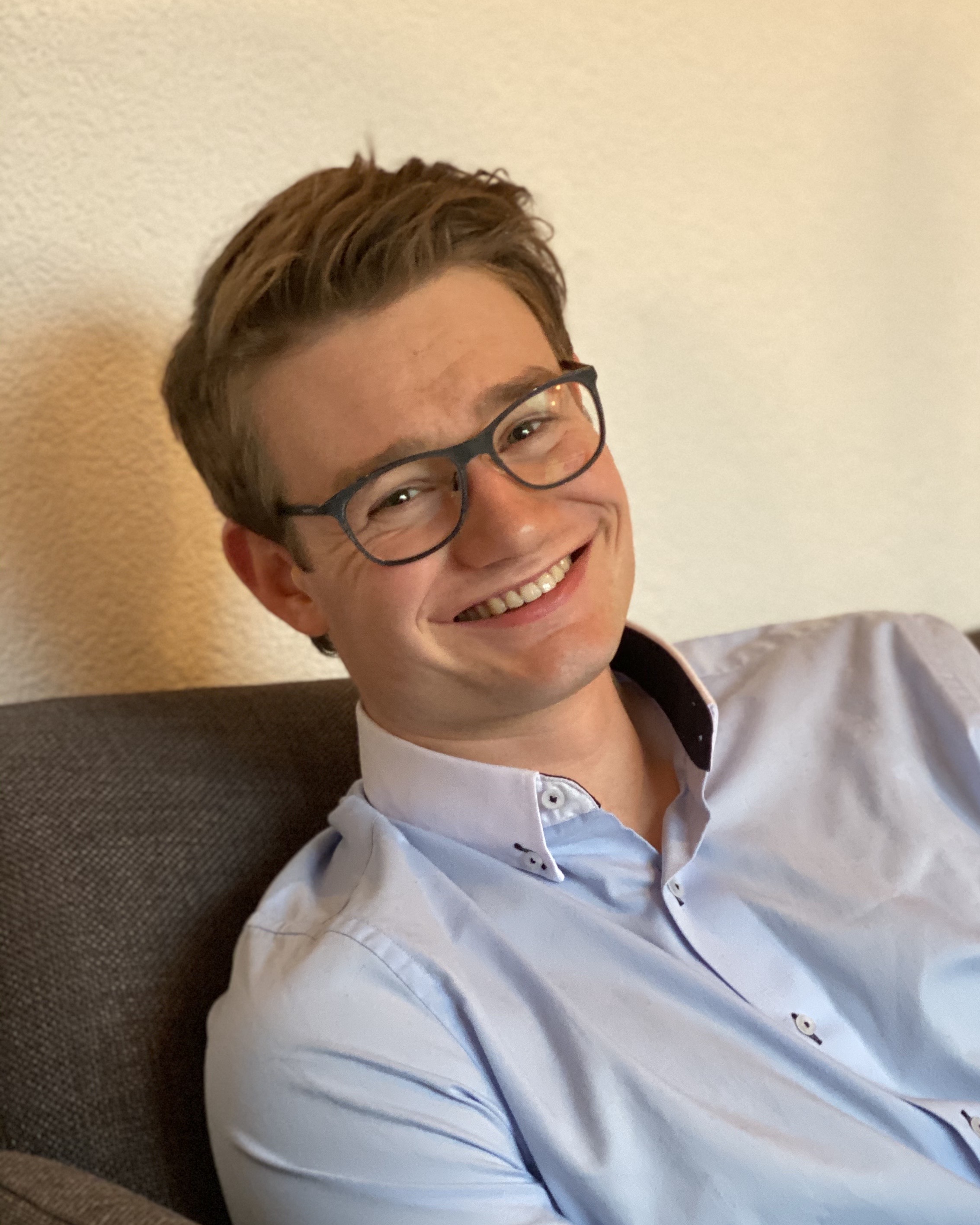}}]
{Julian Moosmann (Graduate Student Member, IEEE)} received the BSc and MSc degrees in electrical engineering and information technologies from ETH Zürich, Zürich, Switzerland, in 2019 and 2023, respectively. From 2022 to 2023, he was a Research Assistant at the Center for Project-Based Learning D-ITET, ETH Z\"urich, Z\"urich, Switzerland, where he is currently conducting his doctorate for pursuing the degree for Doctor of Science with the Integrated Systems Laboratory in conjunction with the Center for Project-Based Learning D-ITET. 
% His research interests include a combination of computer vision, event-based sensing, low-power systems, wireless sensor networks, tiny machine learning / onboard intelligence, and battery-operated distributed systems.
\end{IEEEbiography}

\begin{IEEEbiography}
[{\includegraphics[width=1in,height=1.25in,clip,keepaspectratio]{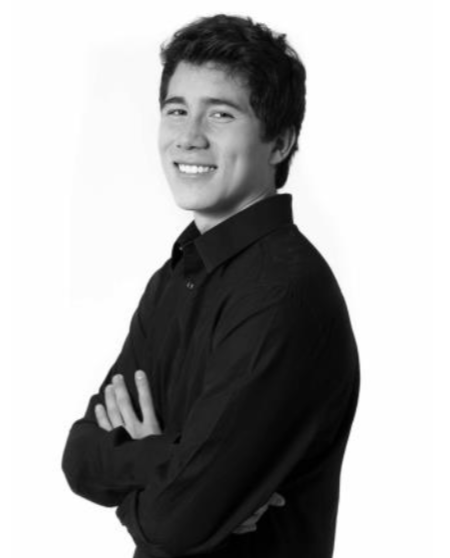}}]{Nicolas Baumann}
was born near Zurich, Switzerland in 1995. He received his BSc and MSc degrees in electrical engineering from ETH Zurich in 2019 and 2022 respectively. Currently, he is a doctoral candidate at the Center for Project-Based Learning at ETH Zurich. His research regards machine learning and sensor fusion in the domain of autonomous driving.
\end{IEEEbiography}

% \begin{IEEEbiography}[{\includegraphics[width=1in,height=1.25in,clip,keepaspectratio]{}}]
% {Seonyeong Heo} received the B.S. and Ph.D. degrees in computer science and engineering from the Pohang University of Science and Technology, Pohang, South Korea, in 2016 and 2021, respectively. Formerly, she was a postdoctoral researcher in the Department of Information Technology and Electrical Engineering, ETH Zurich, Zurich, Switzerland. She is currently an assistant professor in the School of Computing, Kyung Hee University, Yongin, South Korea. Her research interests include real-time embedded systems and tiny machine learning for embedded systems.
% \end{IEEEbiography}

\begin{IEEEbiography}[{\includegraphics[width=1in,height=1.25in,clip,keepaspectratio]{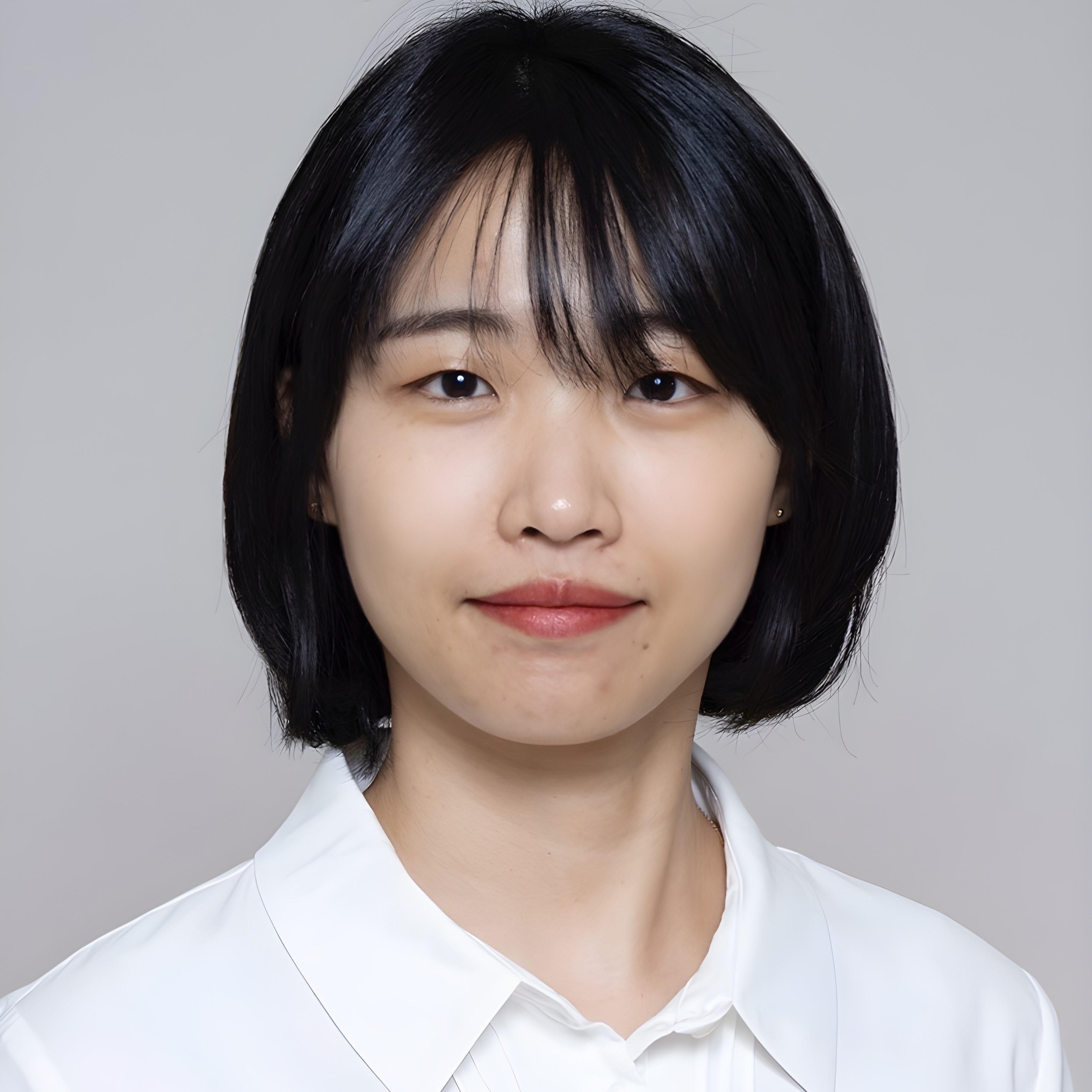}}]
{Seonyeong Heo (Member, IEEE)} received the B.S. and Ph.D. degrees in computer science and engineering from the Pohang University of Science and Technology, Pohang, South Korea, in 2016 and 2021, respectively. Formerly, she was a postdoctoral researcher in the Department of Information Technology and Electrical Engineering, ETH Zurich, Zurich, Switzerland. She is currently an assistant professor in the School of Computing, Kyung Hee University, Yongin, South Korea. Her research interests include real-time embedded systems and tiny machine learning for embedded systems.
\end{IEEEbiography}

\begin{IEEEbiography}
[{\includegraphics[width=1in,height=1.25in,clip,keepaspectratio]{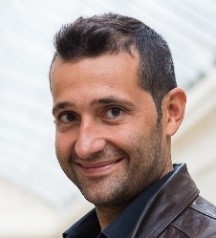}}]{Michele Magno (Senior Member, IEEE)}
 received his master's and Ph.D. degrees in electronic engineering from the University of Bologna, Bologna, Italy, in 2004 and 2010, respectively. \\
Currently, he is a Senior Researcher at ETH Zurich, Zurich, Switzerland, where he is the Head of the Project-Based Learning Center. He has collaborated with several universities and research centers, such as Mid University Sweden, where he is a Guest Full Professor. %He has published more than 150 articles in international journals and conferences, in which he got multiple best paper and best poster awards. The key topics of his research are wireless sensor networks, wearable devices, machine learning at the edge, energy harvesting, power management techniques, and extended lifetime of battery-operated devices.
\end{IEEEbiography}

\end{document}

%% file: abr.tex
\newacronym{mpc}{MPC}{Model Predictive Control}
\newacronym{mpcc}{MPCC}{Model Predictive Contouring Controller}
\newacronym{rl}{RL}{Reinforcement Learning}
\newacronym{mlp}{MLP}{Multilayer Perceptron}
\newacronym{forl}{FoRL}{Foundations of Reinforcement Learning}
\newacronym{ml}{ML}{Machine Learning}
\newacronym{sb3}{SB3}{Stable Baselines 3}
\newacronym{sac}{SAC}{Soft Actor Critic}
\newacronym{ppo}{PPO}{Proximal Policy Optimization}
\newacronym{ai}{AI}{Artificial Intelligence}
\newacronym{nn}{NN}{Neural Network}
\newacronym{sota}{SotA}{State-of-the-Art}
\newacronym{esc}{ESC}{Electronic Speed Controller}
\newacronym{ros}{ROS}{Robot Operating System}
\newacronym{imu}{IMU}{Inertial Measurement Unit}
\newacronym{ekf}{EKF}{Extended Kalman Filter}
\newacronym{slam}{SLAM}{Simultaneous Localization And Mapping}
\newacronym{sdc}{SDC}{Self Driving Cars}
\newacronym{obc}{OBC}{On Board Computer}
\newacronym{qp}{QP}{Quadratic Programming}
\newacronym{uav}{UAV}{Unmanned Aerial Vehicles}
\newacronym{cg}{CG}{Center of Gravity}
\newacronym{em}{EM}{Expectation Maximization}
\newacronym{rms}{RMS}{Root Mean Square}
\newacronym{map}{MAP}{Model- and Acceleration-based Pursuit}
\newacronym{pd}{PD}{Proportional-Derivative}
\newacronym{lut}{LUT}{Lookup Table}
\newacronym{RGB}{RGB}{Red Green Blue}
\newacronym{HYP}{HYP}{Hyper Spectral}
\newacronym{shm}{SHM}{Structural Health Monitoring}
\newacronym{iot}{IoT}{Internet of Things}
\newacronym{tn}{TN}{True Negative}
\newacronym{fp}{FP}{False Positive}
\newacronym{fn}{FN}{False Negative}
\newacronym{mcc}{MCC}{Matthews Correlation Coefficient}
\newacronym{pca}{PCA}{Principal Component Analysis}
\newacronym{sne}{SNE}{Stochastic Neighbor Embedding}
\newacronym{umap}{UMAP}{Uniform Manifold Approximation and Projection}
\newacronym{fomo}{FOMO}{Faster Objects More Objects}
\newacronym{tflm}{TFLM}{TensorFlow Lite Micro}
\newacronym{nba}{NBA}{Normalized Bounding Box Area}
\newacronym{yolo}{YOLO}{You Only Look Once}
\newacronym{mae}{MAE}{Mean Average Error}
\newacronym{cnn}{CNN}{Convolutional Neural Network}
\newacronym{tinyml}{TinyML}{Tiny Machine Learning}
\newacronym{iou}{IoU}{Intersection over Union}
\newacronym{soc}{SoC}{System on Chip}
\newacronym{ram}{RAM}{Random Access Memory}
\newacronym{fps}{FPS}{frames per seconds}
\newacronym{mcu}{MCU}{Microcontroller Unit}
\newacronym{rbe}{RBE}{Reconfigurable Binary Engine}
\newacronym{risc-v}{RISC-V}{Reduced Instruction Set Computer}